# Deep Learning-Based Computer Vision for Real-Time Intravenous Drip Infusion Monitoring

Nicola Giaquinto, *Member, IEEE*, Marco Scarpetta, *Graduate Student Member, IEEE*, Maurizio Spadavecchia, *Member, IEEE*, and Gregorio Andria, *Member, IEEE*

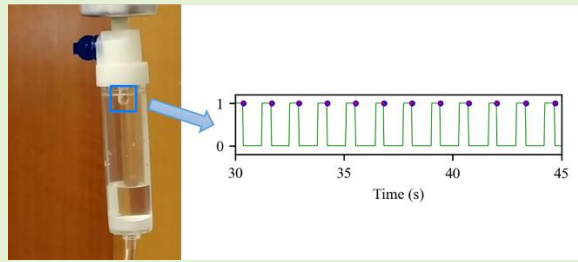

*Abstract*—This paper explores the use of deep learning-based computer vision for real-time monitoring of the flow in intravenous (IV) infusions. IV infusions are among the most common therapies in hospitalized patients and, given that both over-infusion and under-infusion can cause severe damages, monitoring the flow rate of the fluid being administered to patients is very important for their safety. The proposed system uses a camera to film the IV drip infusion kit and a deep learning-based algorithm to classify acquired frames into two different states: frames with a drop that has just begun to take shape and frames with a well-formed drop. The alternation of these two states is used to count drops and derive a measurement of the flow rate of the drip. The usage of a camera as sensing element makes the proposed system safe in medical environments and easier to be integrated into current health facilities. Experimental results are reported in the paper that confirm the accuracy of the system and its capability to produce real-time estimates. The proposed method can be therefore effectively adopted to implement IV infusion monitoring and control systems.

*Index Terms*—healthcare, IV infusion, sensors for therapeutic treatments, drop rate monitoring, biomedical monitoring, deep learning

## I. Introduction

INTRAVENOUS (IV) infusions are one of the fundamental therapies administered in hospitals. They are used to administer medications to patients, to replace lost fluids, to correct chemical or metabolic imbalance, to give fluids to people who cannot drink. The correct operation of IV infusion devices is very important for patients' safety, considering that administering the wrong quantity of the infusion fluid can lead to severe consequences for patients' health. IV infusion sets are therefore frequently checked by manually monitoring the dripping rate that can be observed through their transparent drip chamber. This operation is a burden for medical personnel and like all the human-performed tasks is prone to errors.

Various solutions have been proposed over recent years to deal with automatic IV drip infusion monitoring and control. In [1], a system based on the usage of a pressure sensor to measure the fluid level in the infusion bottle was proposed. The system only requires low-cost hardware and can be used to remotely control multiple infusion sets. The drawback of this method is its invasiveness i.e. the pressure sensor must be in direct contact with the infusion fluid. This is a critical aspect in medical applications, where any possible source of contamination must be avoided.

The Authors are with Politecnico di Bari, Department of Electrical and Information Engineering (DEI), via Orabona 4, I-70125 Bari (BA), Italy (e-mails: N. Giaquinto – nicola.giaquinto@poliba.it, M. Scarpetta – marco.scarpetta@poliba.it,
M. Spadavecchia – maurizio.spadavecchia@poliba.it, G. Andria – gregorio.andria@poliba.it)

The devices proposed in [2], [3] can monitor intravenous infusions by using Microwave Time Domain Reflectometry (TDR) to measure the level of the fluid in the bottle, like had already been successfully done in similar sensing applications [4]. Strip electrodes must be attached on the external surface of infusion bottles. This certainly makes the method safe for medical applications but introduces an additional step to be performed for each infusion bottle, during production or in hospitals.

Drop counting has been used in other works to derivate a measurement of the flow rate of the infusion. Devices proposed in [5]–[13] make use of a light emitter and a light detector, applied on the opposite sides of the drip chamber, to detect the falling of drops and hence to count them. The estimate of the flow rate obtained this way is less accurate than that obtained from a level measurement, since the volume of the drops is only approximately known and can change over time. This specific application, anyway, does not require a highly accurate estimate, and, actually, counting drops is what nurses do to monitor infusions. In [14], an additional light emitter-detector sensor is applied on the tube, to make the system emit an alarm when the infusion is complete.

The devices proposed in [15], [16] use drop counting also, but electrodes are used as sensing elements. They are applied on the infusion bottle and on the drip chamber and are used to measure the capacitance between them. This quantity depends on the properties of the infusion fluid and follows an oscillating trend in accordance with the dripping of the fluid. Drop counting and therefore an estimation of the flow rate is performed analyzing this signal. A capacity measurement is also used in [17]; in this case to produce an estimate of the level of the drip chamber liquid. This quantity, together with drop counting done through an infrared emitter-detector sensor, is exploited to monitor the correct operation of the IV drip infusion.

These last methods do not require the direct contact between sensing elements and the infusion fluid, since the former are applied on the external surfaces of the infusion bottle or of the drip chamber, and this assures that no contamination can occur. Anyway, a non-trivial operation is required to place the sensing elements on the IV infusion kit, that would make it difficult their usage by medical personnel. Introducing the sensing elements during the production of drip infusion kits would be a more viable solution but would require altering the current production processes.

In this paper, the usage of deep learning-based computer vision techniques is explored for the purpose of real-time monitoring of the flow rate in IV drip infusions. This work develops the idea illustrated in [18], i.e. to use a camera to film the drip chamber and analyze the acquired video to detect and count drops, thus obtaining a measurement of the flow rate of the fluid being administered. In [18], a general-purpose object detection neural network was used to process the videos of the drip chamber. The main improvement of this paper is the introduction of a novel neural network that was specifically designed for this application, and is therefore more performant and efficient than the one used in [18]. An objective comparison between the performance of the novel system and the one described in [18] is presented in Subsection III.A.

Cameras are nowadays quite unexpensive devices, easily connected over internet, and therefore very suitable for implementing a centralized monitoring of many IV infusions. The usage of wireless networks to efficiently transmit the images acquired by the camera, that would make the system flexible and easily installable in pre-existent health facilities, is certainly possible considering the recent advances in research in this field [19], [20].

There are other advantages in using cameras: they cannot contaminate the administered fluid (i.e., they are safe by design for medical usage) and do not require other operations beyond the initial installation, so the daily routines of nurses do not undergo any changes as well as any change is imposed to the available IV drip infusion kits.

The proposed device falls into a category of sensors for healthcare that can improve the safety and quality of therapeutic treatments administered to patients. In a typical scenario, the proposed device is installed in each patient's room and sends data to a remote monitoring location, that can be the nurses station of the hospital. Another possible application is in telemedicine programs, where the device can be used to monitor patients at their homes. In this case, the camera can serve two purposes: check whether the IV drip infusion is working correctly, but also monitor the patient from a remote location, e.g. the hospital where they are being treated.

## II. Description of the Flow Rate Estimation Method

An IV drip infusion kit is composed of a bottle containing the infusion fluid, a drip chamber, and a tube connected to the venous catheter. The proposed system uses a camera to film the drip chamber, as shown schematically in Fig. 1. The video acquired by the camera is sent to a neural network that is able to locate the position of the drop and, more important, its state, i.e. whether the drop has just begun to take shape, or is instead well-formed. The alternation of these two states can be easily exploited to count drops and, hence, measure a flow rate.

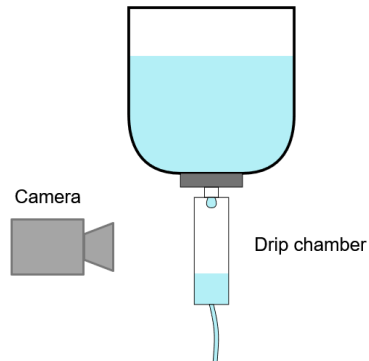

Fig. 1. Representation of the proposed solution

It was decided to use deep learning-based computer vision techniques to process the video acquired by the camera because these videos can have very different features due to e.g. variability of the background texture, of the illumination conditions, of the focus of the camera. This variability make it difficult to use classical image processing techniques (like the closing and snake algorithms used in [21]) because any specific condition must be considered during the development of the processing algorithms to obtain a sufficiently general method. On the other hand, the same neural network can be trained on a larger dataset to make it work in more general conditions. As an example, the neural network can be trained to work in bad illumination conditions, like those that can be expected during nighttime, when patients are sleeping. Obviously, a light source is required for the system to work, since it is based on computer vision; anyway, considering that the camera must be near the drip chamber, a reasonably dim and focused light beam can be used, without disturbing patients.

### A. Neural Network for Drop State Detection

The scheme of the neural network developed for drops detection and counting is shown in Fig. 2. The input of the neural network, a squared $W \times W$ RGB image, is processed through a sequence of convolutional layers to finally produce a $S \times S \times 2$ matrix. In the actual implementation, $W = 416$, and $S = 26$. The indices of the maximum value of this matrix gives the approximate position (first two indices) and the state (third index) of the drop in the input image. Thus, basically, the output of the network is a label containing information about the position and state the drop, encoded as a one-hot matrix (tensor).

The two possible states defined for the drop, i.e.:
- the drop has just begun to take shape
- the drop is well-formed

will be represented, respectively, by the values 0 and 1 of the symbol $s$.

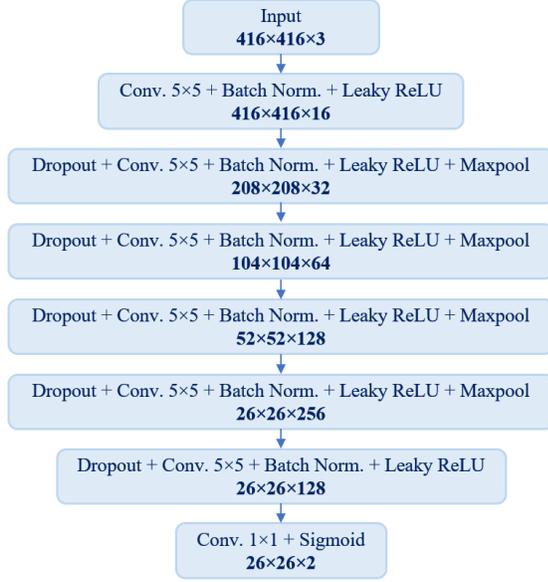

Fig. 2. Architecture of the neural network used for drops detection and counting

The core of the network is a sequence of convolutional layers that use an increasing number of $5 \times 5$ filters and the leaky rectified linear activation function defined as:

$$\phi(x) = \begin{cases} x, & \text{if } x > 0 \\ 0.1x, & \text{if } x \leq 0 \end{cases}$$

These layers are introduced to extract features from the image, with an increasing level of abstraction.

Max pooling, with a stride of 2 between pooling regions and a pooling region size of $2 \times 2$, was applied to the outputs of the central convolutional layers to make the process of features extraction independent on translations in the input and, therefore, to make the model capable of recognizing the drop in any location of the input image.

As shown in Fig. 2, both Dropout [22], with a 0.2 drop rate, and Batch normalization [23] were used in the network. This choice proved very effective in improving the convergence of the model, avoiding overfitting.

The last layer of the network classifies the features extracted by the previous layers and organizes them in the $S \times S \times 2$ output matrix. It uses $1 \times 1$ filters for convolutions and the sigmoid activation function:

$$\phi(x) = \frac{1}{1 + e^{-x}}$$

which guarantees that the elements of the output matrix are between 0 and 1.

Overall, the designed neural network model counted about 1.9M parameters.

The proposed neural network can be used to simultaneously analyze multiple streaming videos also. Let $N_S$ be the number of input video streams. $N_F$ frames per camera must be acquired and stacked in a $N \times W \times W \times 3$ tensor, where $N = N_S N_F$. This tensor can be directly analyzed by the neural network in a more efficient way than analyzing a single frame at a time, since the parallel computing capabilities of Graphics Processing Units (GPU) can be fully exploited this way. The output of the network would be a $N \times S \times S \times 2$ tensor, that could be easily unpacked to obtain $S \times S \times 2$ tensors like those described before. More details on processing multiple streams are given in Subsection III.A.

### B. Dataset for Training the Neural Network

The dataset used for training the neural network was created starting from a set of videos of an operating IV drip infusion kit. While filming these videos, the infusion kit was regulated to produce a low drop rate so that a great number of contiguous frames containing the drop in a certain state was available. This was useful for quickly labeling the frames, i.e. specifying the position and the state of the drop in each frame. In fact, regions of frames with the drop in both states were selected in each video and, in each of these regions, the position of the drop was marked. Given that the drip chamber was not moving with respect to the camera, it was necessary to mark the position of the drop in one frame per region only, to be able to extract a great number of labeled images from the videos.

Subsequently, the extracted frames were cropped to obtain squared images of size $W \times W$. This operation was performed in such a way that the positions of the drops were equally distributed over the $W \times W$ surface. Besides, uniformly distributed-random zooming levels, between 0.9 and 1.1, were applied to the starting images.

Finally, the label of each sample of the dataset was defined as a $S \times S \times 2$ matrix, whose elements were:

$$Y^{(n)}_{i,j,k} = \begin{cases} 1, & \text{if } i = \left\lfloor \frac{x^{(n)} \cdot S}{W} \right\rfloor, j = \left\lfloor \frac{y^{(n)} \cdot S}{W} \right\rfloor, k = s \\ 0, & \text{otherwise} \end{cases}$$

where $n$ is the index of the sample, $[x^{(n)}, y^{(n)}]$ is the position of the drop in the sample ($0 \leq x^{(n)} < W$, $0 \leq y^{(n)} < W$), $s$ is state of the drop.

The dataset obtained as described above was composed of about 15k samples, equally distributed between the two classes. Some examples taken from the dataset are reported in Fig. 3, with overlapped a $26 \times 26$ grid representing their label, i.e. the matrix $Y^{(n)}_{i,j,k}$, (the colored cell represents the position and the $k$ value corresponding to the "1" in the matrix). As it can be seen from the figure, the drip infusion set was filmed from different sides, with different backgrounds and illumination conditions, to have a more general representation of the two states of the drop.

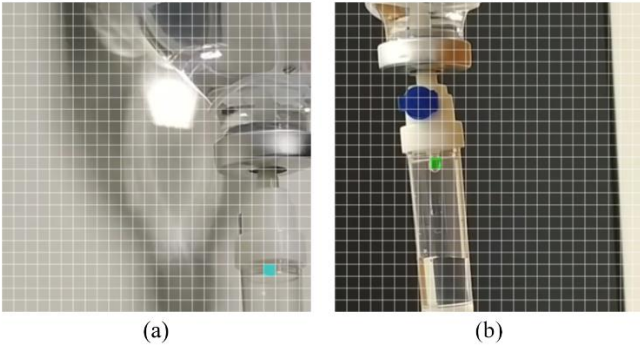

Fig. 3. Samples from the training dataset: drop in the state $s = 0$ (a), drop in the state $s = 1$ (b).

### C. Training of the Neural Network

The training process involved the optimization of the loss function:

$$L(Y^{(n)}, \widehat{Y}^{(n)}) = \sum_{i=0}^{S-1} \sum_{j=0}^{S-1} \sum_{k=0}^{1} Y^{(n)}_{i,j,k} \log \widehat{Y}^{(n)}_{i,j,k}$$

where $Y^{(n)}$ is the label of the n-th sample of the training dataset, $\widehat{Y}^{(n)}$ is the output that the neural network produces when its input is the $n$-th sample of the dataset. This kind of loss function is basically a categorical cross-entropy computed on a matrix instead of a vector, and it was chosen because it is the most effective for one-hot encoded quantities.

The dataset described in the previous paragraph was used to train and validate the neural network, splitting it according an 80/20 ratio (training/validation). It took about five epochs for the model to converge, as can be seen from the loss versus epochs diagram in Fig. 4.

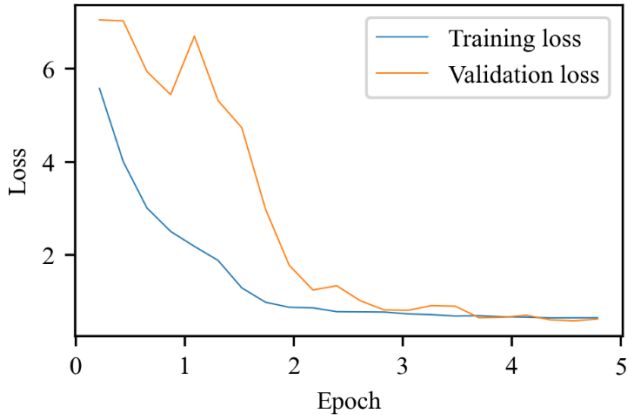

Fig. 4. Loss progress during training.

At the end of the training, the accuracy in the prediction of the state of drops was 100% for the validation dataset. As an example, the output of the neural network for one of these samples is reported in Fig. 5 (in the form of a "heat map"). The second layer of the output matrix, related to the state $s = 1$, is zero everywhere, while in the first layer, related to the state $s = 0$, there are two points near 1. This is due to the drop being at the edge of two cells of the $S \times S$ grid. This is not an issue, since we are not interested in knowing the position of the drop with great accuracy.

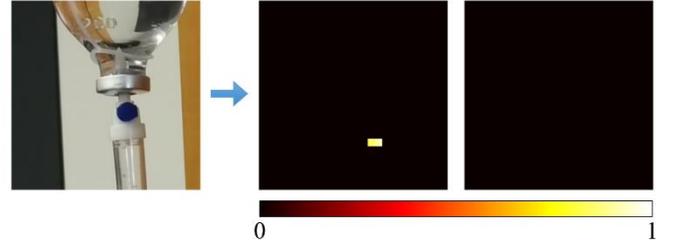

Fig. 5. Output of the neural network for a sample from the validation dataset.

### D. Flow Rate Estimation

The drop counting operation is straightforward, given the output of the neural network for each frame of the video acquired through the camera. The only sufficient information is the measured state of the drop, computed from the output matrix $\widehat{Y}$ as:

$$\hat{s} = \arg\max_{k} \left( \max_{i,j} \widehat{Y}_{i,j,k} \right)$$

The drop count is increased when the state of the drop changes from 1 to 0.

The instantaneous flow rate (in drops per minute, gtt/min) at which the drip infusion kit operates is computed at each time instant $t_i$ in which the drop count is increased, by the formula:

$$Q(t_i) = \frac{N}{t_i - t_{i-N}}$$

where $N \in \mathbb{N}^+$ is a prefixed number of intervals between consecutive drops, to be observed to obtain a measure. It is possible to choose $N = 1$, but in this way the quantity has sometimes an excessive variability, which is an objective characteristic of the flow. The value $N = 3$, chosen empirically, demonstrated to be a good compromise between the responsiveness of the system and the stability and meaningfulness of the measured flow rate. Of course, measurements with a different value for $N$ is easily implemented.

The estimated position of the drop in the image is not taken into account while measuring the flow rate of the IV drip infusion set, but it is useful to detect anomalies, such as an excessive movement of the camera with respect to the bottle. The system is robust with respect to such movements, but it must be guaranteed that the drop remains in the camera framing. For this purpose, an alarm can be emitted on the monitoring interface if the position of the drop approaches the edges of the framing, to inform the operator that a manual repositioning of the camera (or the infusion set) is required. Problems relative to framing can also be completely avoided by designing a specific support for the camera to be applied directly on the infusion bottle e.g. using a clip. Such a solution can be easily engineered also with an embedded light source to illuminate the drip chamber in low ambient light conditions.

## III. EXPERIMENTAL RESULTS

A set of videos with different characteristics was acquired to assess the performance of the system in a plausible working condition. While filming these videos, the IV drip infusion kit was set to produce a rather typical flow rate, between about 20 and 50 gtt/min. Besides, to properly evaluate if the neural network was generalizing well, due precautions were taken to make the test videos different from the videos used to produce the training dataset, i.e. using different backgrounds, illumination conditions and camera framings. The drip chamber, as it appears in the test videos, is depicted in Fig. 6.

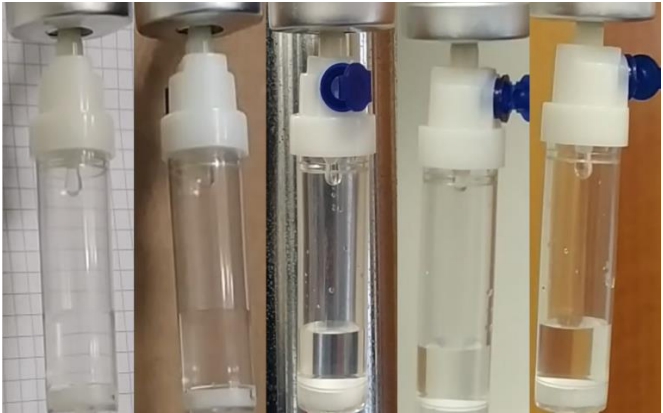

Fig. 6. Images of the drip chamber extracted from the videos used to test the system.

### A. Drop State Estimation

The neural network described in the previous Section was implemented in TensorFlow [24]. After the training process, it was used to analyze the test videos and estimate the evolution with time of the state of the drop. This quantity has been therefore used for computing the instantaneous flow rate of the IV drip infusion kit.

The neural network has been capable of processing about 240 FPS on a NVIDIA Tesla P100 GPU, which means it can handle $N_S = 7$ input streams at 30 FPS, considering a conservative margin. If necessary, a system with more GPUs can be built using common commercially available technologies, to handle more input video streams.

The results of the drop state detection are reported in Fig. 7 for the first 20 s of the five test videos, together with the drop counting results and the true values that were obtained by manually labeling the videos. As it can be seen in the figure, the measured position of time instants in which the drop detaches from the dripper are very accurate. The only visible defects are some fast transitions between the two states in the second video: however, given that they last only for one frame, they were easy to filter out, resulting in a correct drop count.

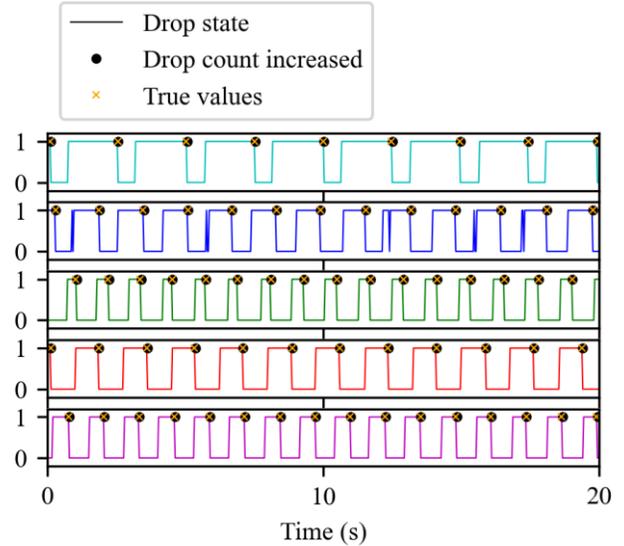

Fig. 7. Detected drop state versus time, and instants in which the drop count is increased, for portions of five different videos taken in the five conditions represented in Fig. 6. "True values" are the instants in which the drop detaches from the dripper, determined by human inspection of the processed videos.

These results were compared with those previously obtained by the Authors in [18], a preliminary work in which an existing neural network model for object detection, YOLO [25], was used to distinguish frames with a drop from frames without any drops. In Fig. 8 the results of drop counting obtained with the two methods are reported. The method proposed in this paper is a clear improvement over [18] that, even if able to produce good final estimates, required more elaborated post-processing operations to compute the flow rate, and was more prone to counting errors.

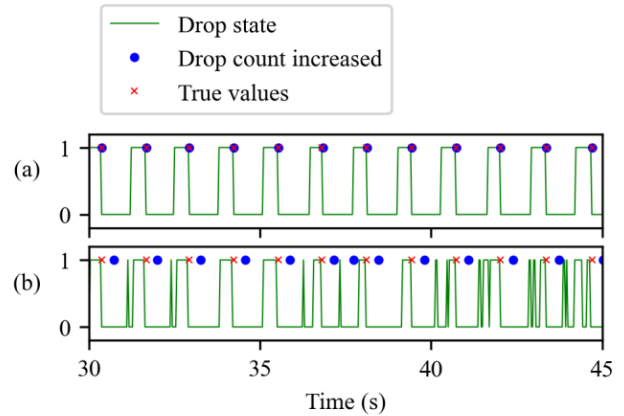

Fig. 8. Results of drop counting for a region of the third video, using the method proposed in this paper (a), and using the method in [18] (b).

A comparison between the obtained results and other drop-counting solutions is appropriate. The most popular solution in the literature is based on photo-detection. Such systems are generally very accurate. In [6] a mean error of 1.65 gtt/min is reported; in [9] a maximum percentage error of 4.3% over the total number of drops counted in tests was observed; and a 100% accuracy in drop counting is reported in [17]. The system presented here has counted 100% of the drops in the test set, with zero false positives, and has exhibited therefore a 100% accuracy, the same obtained in [17], without designing a specific hardware, firmware or arranging a complex testbed.

*B. Flow Rate Estimation*

The final output of the proposed method is the estimate of the flow rate of the IV drip infusion kit reported in Fig. 9 for the five test videos. The results show that the system is able to produce a very accurate estimate of the flow rate.

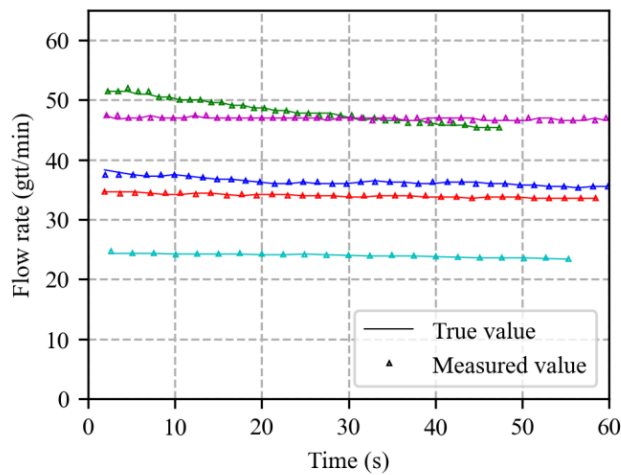

Fig. 9. Estimation of the flow rate of the IV drip infusion kit in the five test videos. The "true value" of the flow rate is computed by human drop counting.

## IV. CONCLUSIONS

In this paper, a deep learning-based computer vision method for IV drip infusion monitoring and flow rate measurement has been proposed. The system has been conceived for easy remote monitoring of traditional IV infusion systems. The advantages of this method are the following:
- it does not interfere with the normal operations required for traditional IV infusion (simple installation process, no risk of contaminating the IV fluid, few interferences with the work of medical staff);
- it counts the drops virtually without error;
- it is easily usable to monitor automatically and simultaneously several different IV infusions, in different locations, provided an internet connection is available.
- it is, finally, useful to avoid unnecessary risks while monitoring contagious patients in isolations.

A further development of the system could also involve the estimation of the volume of the drops, in order to obtain a flow measurement e.g. in milligrams per minute, and to monitor the amount of fluid administered over time.

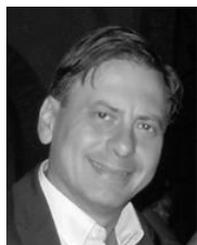

**Nicola Giaquinto** (M'12) received the M.S. and Ph.D. degrees in electronic engineering from the Polytechnic University of Bari, Bari, Italy, in 1992 and 1997, respectively.

From 1997 to 1998, he was a Post-Doctoral Researcher with ENEA (Italian Agency for New Technologies), Casaccia Research Center, Rome, Italy. In 1998, he joined the Polytechnic University of Bari, where he has been a Coordinator of the Measurement Laboratory, an Assistant Professor, and an Associate Professor since 2004. Since 2013, he has been the Dean of Internationalization (Student and Staff Mobility Programs), and since 2018, he has been a member of the Academic Senate, Polytechnic University of Bari. His research interests are focused on mathematical methods for measurements (statistical signal processing, system identification, and AI techniques for measurement applications), theoretical and practical issues in measurement uncertainty evaluation, metrology of waveform recorders and signal generators (A-to-D and D-to-A converters), measurement system analysis, design and characterization of sensors and measurement methods (e.g., reflectometry-based distributed sensing).

Dr. Giaquinto was a three-time winner of the Best Reviewer of the Year recognition from the IEEE Instrumentation and Measurement Society.

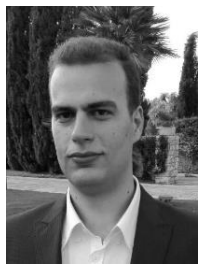

**Marco Scarpetta** received the M.S. degree in electronic engineering from the Polytechnic University of Bari, Bari, Italy, in 2018, where he is currently pursuing the Ph.D. degree.

Since 2019, he has been with the Electrical and Electronic Measurements Group, Polytechnic University of Bari. His current research interests are in the field of electrical and electronic measurements, signal processing, and artificial intelligence techniques.

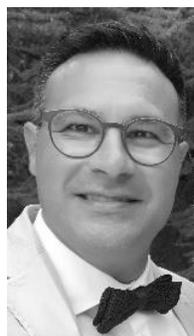

**Maurizio Spadavecchia** (M'14) received the M.S. (Hons.) and Ph.D. degrees in electrical engineering from the Polytechnic University of Bari, Bari, Italy, in 2006 and 2011, respectively.

Since 2006, he joined the Electrical and Electronic Measurements Laboratory, Polytechnic University of Bari, where he is currently a Post-Doctoral Researcher. His current research interests include signal processing, the characterization of different kinds of instruments and devices, the characterization of renewable energy devices, software for automatic test equipment, photovoltaic and thermoelectric modules' modeling and testing, power quality systems and measurements, smart metering, and soil mechanics testing equipment.

Dr. Spadavecchia is a member of the IEEE Instrumentation and Measurement Society and the Italian Association Electrical and Electronic Measurements Group.

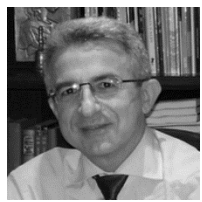

**Gregorio Andria** received the M.S. degree in electrical engineering and the Ph.D. degree in power systems from the State University of Bari, Bari, Italy, in 1981 and 1987, respectively. He is currently a Full Professor of electrical measurements with the Polytechnic of Bari, Bari, where he is the Dean of the II Faculty of Engineering. His research interests are focused on the optimization of spectral estimation algorithms for monitoring power systems and diagnostics of electrical drives, design of digital filters for digital instrumentation, characterization of A/D converters, digital signal processing, and characterization of intelligent instruments. Dr. Andria is a member of the Italian Group of Electrical and Electronic Measurements (GMEE).